\documentclass[10pt, a4paper]{article}
\usepackage{lrec2022} 
\usepackage{multibib}
\newcites{languageresource}{Language Resources}
\usepackage{graphicx}
\usepackage{tabularx}
\usepackage{soul}
\usepackage{amsmath}
\usepackage{titlesec}

\titleformat{\section}{\normalfont\large\bf\center}{\thesection.}{1em}{}
\titleformat{\section}{\normalfont\large\bfseries\center}{\thesection.}{1em}{}
\titleformat{\subsection}{\normalfont\SmallTitleFont\bfseries\raggedright}{\thesubsection.}{1em}{}
\titleformat{\subsubsection}{\normalfont\normalsize\bfseries\raggedright}{\thesubsubsection.}{1em}{}
\renewcommand\thesection{\arabic{section}}
\renewcommand\thesubsection{\thesection.\arabic{subsection}}
\renewcommand\thesubsubsection{\thesubsection.\arabic{subsubsection}}

\usepackage{epstopdf}
\usepackage[utf8]{inputenc}

\usepackage{hyperref}
\usepackage{xstring}

\usepackage{color}

\title{Russian Jeopardy! Data Set for Question-Answering Systems}

\name{Elena Mikhalkova, Alexander Khlyupin} 

\address{Tyumen State University \\
         625003, Volodarskogo, 6, Tyumen, Russia \\
         e.v.mikhalkova@utmn.ru \\}

\abstract{
Question answering (QA) is one of the most common NLP tasks that relates to named entity recognition, fact extraction, semantic search and some other fields. In industry, it is much valued in chat-bots and corporate information systems. It is also a challenging task that attracted the attention of a very general audience at the quiz show Jeopardy! In this article we describe a Jeopardy!-like Russian QA data set collected from the official Russian quiz database Chgk {\em che-ge-`ka:}. The data set includes 379,284 quiz-like questions with 29,375 from the Russian analogue of Jeopardy! -- ``Own Game''. We observe its linguistic features and the related QA-task. We conclude about perspectives of a QA challenge based on the collected data set.
\\ \newline  }

\begin{document}

\maketitleabstract

\section{Introduction}
In natural language processing (NLP), question answering (QA) is one of the most common tasks that encompasses a number of question types, including ``questions about everything'', the so-called open-domain QA~\cite{chen2020open}. Open-domain questions cover a wide range of topics and do not necessarily come in form of an actual question (e.g. ``Who is the living Queen of England?'') which draws the task of answering them very close to information retrieval. The query can be just a line of keywords: {\em living Queen England}, but pragmatically it is still a question. From this broad perspective, QA is developed in production of search engines, corporate information systems and conversational technologies like chat-bots.

In February 2011, Watson, an IBM's information system~\cite{ferrucci2010building} installed in a small computer, won against two very prominent human players in a TV quiz-show called Jeopardy!~\footnote{\url{https://www.nytimes.com/2011/02/17/science/17jeopardy-watson.html}} The algorithm was trained on TREC corpus~\cite{voorhees1999} and 500 questions manually collected from the TV-show. In TREC, questions are formulated quite typically, e.g. ``How many calories are there in a Big Mac?'', although they cover a variety of topics. In contrast to it, the Jeopardy! challenge presents questions as clues narrowed by a certain domain like in the following example from~\cite{ferrucci2010building}:

{\em Category:} Oooh….Chess

{\em Clue:} Invented in the 1500s to speed up the game, this maneuver involves two pieces of the same color.

{\em Answer:} Castling

The existing open-source Russian QA data sets are more like trivia questions and answers resembling TREC: RuBQ~\cite{korablinov2020rubq} consists of 1,500 Russian questions loaded from various ``quiz collections on the Web'' with answers linked to Wikidata entities; RuBQ 2.0 has ``2,910 questions along with the answers and SPARQL queries''~\cite{rybin2021rubq}; SberQuAD~\cite{efimov2020sberquad} contains ``50,364 paragraph–question–answer triples'' that are now publicly available; the questions were written by crowd-annotators.

In this article, we observe a data set of Russian Jeopardy! questions and answers and outline a related QA challenge. The database of questions and answers called ChGK {\em che-ge-`ka:} is freely available at \url{https://db.chgk.info/}. Our current contribution includes the following:

\begin{enumerate}
    \item We describe the Russian Chgk QA database containing nearly 400K questions and answers that test players' logic and erudition.
    \item We describe its sub-corpus of Jeopardy!-like questions and outline its characteristic features.
    \item We formulate a QA-challenge based on the Russian Jeopardy! data set.
\end{enumerate}

\section{Russian ChGK Database}

There exists a variety of Russian intellectual games (quizzes) some of which have formed very devoted communities in and even outside Russia. ``What? Where? When?'' ({\em Chto? Gde? Kogda?}, hence the abbreviation Ch-G-K) is one of the most popular Russian TV quiz shows, dating back to 1975~\footnote{\url{https://en.wikipedia.org/wiki/What\%3F_Where\%3F_When\%3F}}. As the TV game show allows but a few players (a team of six) per one episode, in the 1990s the game format spread among common people who wrote questions and played them at local ChGK tournaments. The movement grew into the so-called ``Sports Chgk''. The community site that collects information about the movement~\footnote{\url{https://rating.chgk.info/}} contains ratings of 228,438 players from 54,995 teams (as of 7 May 2022). ChGK tournaments are organized in Montreal, Richmond Hill, Vilnius, Odessa, 
Kharkiv, Cologne, Boston, Nahariya, Eilat, Parnu, Astana, Vladivostok and many other cities all over the world. The movement has an official open access collection of about 400K questions in Russian. In view of its size, metadata, effort of the community that supports it, this database can be considered cultural heritage of the Russian language. The earliest tournament in the database is the 1990's ``I Championship MAK in ``What? Where? When?'' 1990-01-01''.~\footnote{\url{https://db.chgk.info/tour/mak1}} The copyright allows to use it for non-commercial purposes with some of the packs (collections of questions played during one tournament) distributed under different Creative Commons licenses~\footnote{\url{https://db.chgk.info/copyright}}. Packs are written, tested, approved and then played at different events (offline and online) under different commercial and non-commercial terms. After a row of tournaments, packs are uploaded to the database. Amateur and semi-professional packs usually do not go to the official database. The moment the packs are uploaded they are under the database license.

Due to the number of people involved in the movement and not only for sports, but commercial interest, we can say that for some people sports ChGK is a profession. Our experience of communicating with the community shows that writing questions for ChGK is demanding and depends on authors' reputation. There are well-known authors who earn money by writing and testing questions, and entrepreneurs who organize commercial tournaments. Hence, we can say that the ChGK database contains {\em professionally written} questions, maybe, not from the beginning of the 1990s, but from the times when it became business for many authors and organizers.

\begin{figure}[!h]
\begin{center}
\includegraphics[scale=0.33]{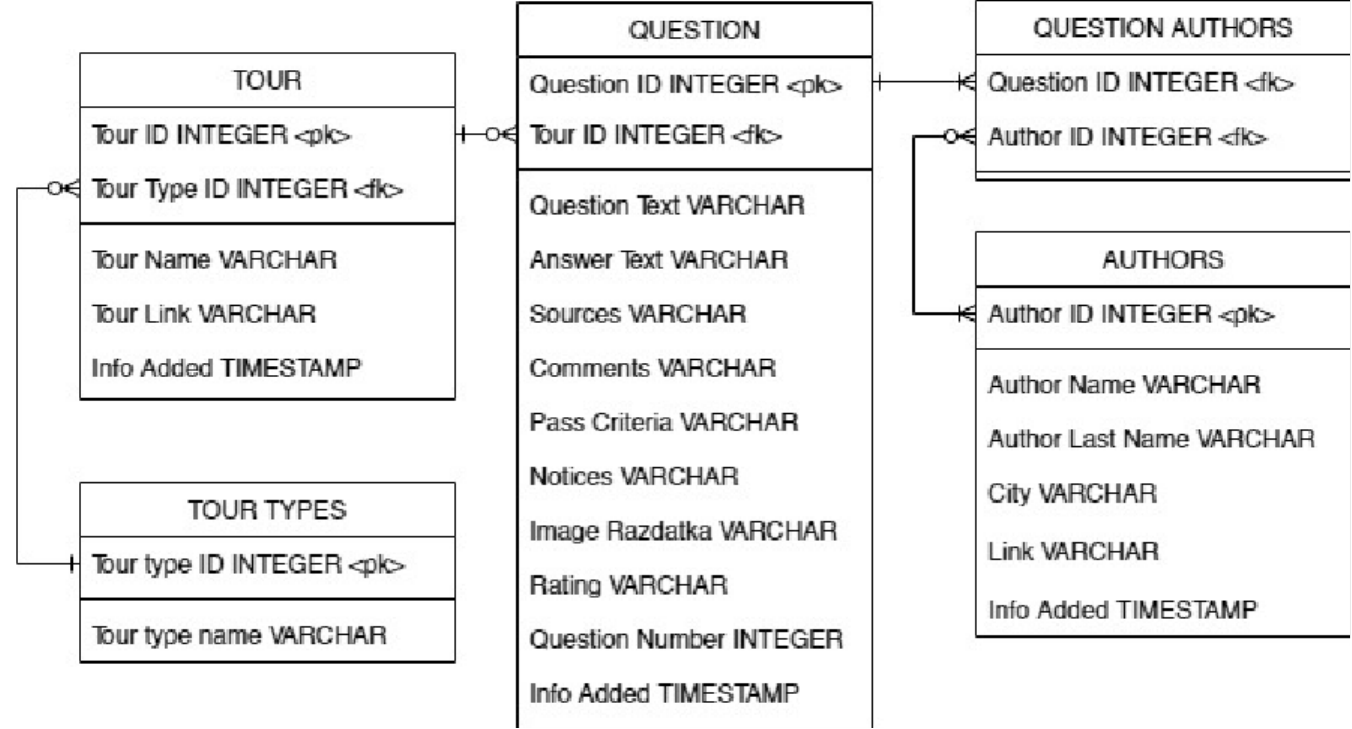}
\caption{A part of our downloaded ChGK database illustrating metadata about questions.}
\label{fig:DB}
\end{center}
\end{figure}

As the website allows only specific search in the database, we parsed the XML-tree of tournaments at \url{https://db.chgk.info/tour} with the Python library BeautifulSoup~\footnote{\url{https://www.crummy.com/software/BeautifulSoup/bs4/doc/\#}} and gathered all the QA information from HTML-pages. The general metadata include: Question, Answer, Author, Sources (Web-links that authors used to write a question), Comments (by authors and organizers), Pass Criteria (in case players' answers are not very precise), Notices (comments by players), Images (Web-links to pictures if they are needed in a question), Rating (hardness of the question calculated from how many teams managed to answer it), Number of question (in order in each pack), Tournament type. The metadata for Jeopardy! questions also include Topic (a common topic for a set of 5 questions, traditionally called ``a category'') and Topic Number (in the order of sets of questions from one tournament). The data were collected in form of .csv tables locally and uploaded to our own SQL-database (see a part of its scheme referring to the corpus in Fig.~\ref{fig:DB}).

As mentioned, the sports ChGK includes questions of different types depending on the type of tournament that they are played at. In total, there are nine types: author's, championships at different countries and regional tournaments (the format can be purely of its authors' design, although it usually complies with the general style), ``synchrons'' (typical ChGK questions played at tournaments simultaneously by many teams), Internet and television quizzes, questions for training, topical questions, questions in a poetic form (verses), questions for erudition (i.e. based purely on knowledge) and of the Jeopardy! type. The most of the database resembles the following example.

\textbf{Question 7, the tournament ``ChGK is... - 2017'':}~\footnote{Authors V. Ostrovskiy, A. Boyko, M. Podryadchikova. Translation into English is ours. \url{https://db.chgk.info/tour/eila08al.2}} The legend has it that once Paul Bunyan fired his gun at a deer and ran to get his prey. But he ran so fast that he DID THIS and felt an itch in his back. What did he do?

\textbf{Answer:} He outran the bullet.

This question was played at tournaments for teams of two players, usually held on the 14 of February. Hence, its title, that resembles the name of the chewing gum ``Love is..''. It is supposed to be an easy question so that a small team can solve it; experienced players would consider it straightforward, giving out a lot of details to find the correct answer. Questions in tournaments for teams of six are a lot more obscure: they give too many or too few details, contain misleading metaphors.

The question is written according to a very common formula: ``DID THIS''. Often the question-like part (``What did he do?'' in the example above) is omitted. At tournaments, the host reading it would put an additional stress on the phrase ``DID THIS'', so that the players understand that this is a question in form of a statement. In the database, such phrases are italicized or capitalized like in our example.

Although the question mentions the detail -- the American folk hero Paul Bunyan and a deer -- the answer does not require this information. It can be derived only from the situation with the bullet, running and an itch in the back. This is a way of misleading players that grab at several hints and need to choose the correct semantic, logical and factual track that narrows the choice of answer. It is important that answers derived from wrong tracks should be incoherent or contradict some facts that are omitted in the question, so that the correct answer cannot be criticized. In the classification offered by~\cite{dimitrakis2020survey}~\footnote{Which they attribute to~\cite{mishra2016survey}, but the classification by \newcite{mishra2016survey} is only a part of their list.}, such questions are called {\em procedural}.

The ChGK database also contains questions of Jeopardy! type. In Russian, Jeopardy! is called ``Own Game'' {\em Svoya Igra}. ~\footnote{\url{https://www.imdb.com/title/tt1381017/}} Nearly all these questions are in form of statements and the object of interest, about which the question is asked, is often capitalized. Let us study the following example.

\textbf{Topic 7: Parrots. Question No. 3.}~\footnote{Author Oleg Sarayev. Translation into English is ours. \url{https://db.chgk.info/tour/eu05stsv}} The last name of THIS famous DETECTIVE is translated as ``a parrot''.

\textbf{Answer:} Hercule Poirot.

Note that this question is shorter and more fact-oriented than the ChGK question before it. It is meant for single players competing against each other, although there are variants of ``Own Game'' for teams of two, three and four. The question above requires to compare two rows of data: words denoting ``parrot'' in different languages and last names of famous detectives. In the classification by~\cite{dimitrakis2020survey}, this type of questions is called factoid, meaning that it resembles a fact, but it is missing some information. Some ChGK questions are also very close to this type, especially in tournaments called {\em lite}, i.e. easier tournaments for new-comers and younger players.

Like in Jeopardy!, questions in ``Own Game'' are organized according to topics (categories) in packs of five, from the easiest to the hardest. The number of the question in the example above is 3 which denotes that it is of medium hardness. At the sports ``Own Game'', i.e. not the television version, the player who answers it faster than others will get 30 points (the easiest question weighs 10 points and the hardest -- 50).
 
Often the topic is a direct hint to the answer, so adding topics to a QA system's input is supposed to be useful. Consider the following question:

\textbf{Topic:} OST. \textbf{Question:} Van Gogh, Gauguin, and Toulouse-Lautrec belonged to THIS movement. \textbf{Answer:} Post-impressionism.~\footnote{Author Yuri Grishov. Translation into English is ours. \url{https://db.chgk.info/tour/grishov}}

``OST'' means that this combination of letters should be in the answer, which leads to just ``impressionism'' not being the answer. In the following case, it is impossible to derive the question without knowing the topic:

\textbf{Topic:} Authors of questions. \textbf{Question:} Are you jealous? \textbf{Answer:} Paul Gauguin.~\footnote{Author Yuri Grishov. Translation into English is ours. \url{https://db.chgk.info/tour/grishov}}

The topic ``Authors of questions'' presupposes that players should remember famously known questions like the one which is the title of Gauguin's picture.

Although usually fact-oriented, the Russian Jeopardy! questions can be of the logical type as well:

\textbf{Topic:} Don Aminado. \textbf{Question:} Of the two who are going to bet, the both risk: one -- to lose, the other -- ...  \textbf{Answer:} To never be paid.~\footnote{Author Yuri Grishov. Translation into English is ours. \url{https://db.chgk.info/tour/grishov}}

The question above is based on a quotation by a famous Russian writer and entails knowledge of a real-life situation, although questions of this type are not very common in the Russian Jeopardy! database. The game does not only check who knows more facts and can recall them faster than others. It also checks how accurate players are when they evaluate possibility that the answer they have just come up with is correct. Rare logical questions based on common sense and typical situations, evidently, aim at the latter skill.

It is also important that some questions in the Chgk database depend on additional media:

\begin{enumerate}
    \item images (but not videos; however, this is not the case with the TV game show) which can not only be photos, screenshots, etc., but visual aids like schemes, symbols, texts printed on handouts (called ``razdatka'' and named so in our scheme~\ref{fig:DB});
    \item the host's intonation. Comments to questions often contain remarks on how to pronounce some parts, for example, without giving out the answer. Intonation can also be marked with capital letters.
\end{enumerate}

\section{Russian Jeopardy! Data Set}

As mentioned, we downloaded data from the official sports ChGK resource to be able to parse them and store in different formats. The Russian Jeopardy! ({\em Own Game}) data set seems to us to be the most valuable for NLP as:

\begin{enumerate}
    \item its questions are shorter than in other quizzes and more fact-oriented;
    \item it is quality-guaranteed, as it was created by professional authors;
    \item it is suitable for open-domain QA;
    \item it has additional information like links to Web-sources and question ranking that points at its ``hardness'';
    \item above all, it is not too trivial in the field of QA data sets and hence it can foster new tasks and approaches in QA itself.
\end{enumerate}

The last point is more vividly discussed by~\cite{boyd-graber-borschinger-2020-question}.

Currently, the data set~\cite{russianjeopardy} contains 29,375 questions from the ChGK database. The questions were selected based on the following criteria:

\begin{enumerate}
    \item a question is in form of a text;
    \item a question does not have an image supporting it;
    \item a question does not mention that any images should be distributed or shown on a screen while solving it.
\end{enumerate}

I.e. these are fully verbalized questions. The data set including some ``flattened'' metadata from the database scheme~\ref{fig:DB} is stored in a .csv file\footnote{\url{https://github.com/evrog/Russian-QA-Jeopardy}}. The delimiter is tabulation. The data include: Question ID, Question, Answer, Topic, Authors' Full Names, Name of tournament, Link to Tournament. The rest of the information supplying questions has not been included in the data set as it is not given to players during the game, but it is available via links to tournaments.

Table~\ref{tab1} gives a summary of the whole ChGK data set, as downloaded on 6 July 2021 and updated on 18 November 2021. In the table, ``Synchron'' is a typical ChGK tournament played immediately by several teams of six players maximum; ``Lite'' is its mentioned version with easier questions. Its questions are shorter, but unlike Jeopardy! they are more logic-oriented. The table demonstrates that Jeopardy! questions are twice shorter in length than typical ChGK questions. And even lite questions are not near them in length.

Table~\ref{tab2} describes distribution of words of different parts of speech across the Jeopardy! data set, classified with the help of NLP-software spaCy~\cite{matthew}. It is of no surprise that nouns are the most frequent category, but, among them, proper nouns stand out. Proper nouns are much less frequent in topics (\~16\%) and questions (\~17\%). However, they are in \~34\% of answers. NER-classifier by spaCy defines that persons, organizations and locations are approximately equally distributed in questions. However, in answers persons comprise as many as 70\% of the classified entities.

It is also natural that verbs are about three times more frequent in questions, than in answers and topics. The most frequent verb is ``to name'' (3,607 tokens), probably, due to a typical formula of a question ``Name somebody or something that..'' which is a variant of ``THIS somebody or something..''. As for other actions expressed by verbs, beside ``being'' or ``becoming'' and their variants, they are ``to gain'' (783 tokens), ``to wear'' (649), ``to write'' (546), ``to have'' (425), ``to be located, situated'' (411), ``to say'' (360), ``to call'' (360), ``to tell'' (348), ``to belong'' (334), ``to mean'' (333), ``to do'' (295), ``to play'' (284), ``to write'' (275), ``to be considered'' (266), ``to happen'' (263), ``to create'' (257), ``to paint, describe'' (242) etc. These verbs hint at a more general topic, for example, art, poetry, music, sports, awards, famous quotes. Although, due to polysemy, the verb ``to gain'' is used in quite a variety of topics.

\begin{table*}[ht]
\begin{center}
\begin{tabular}{|l|l|l|l|l|}
\hline
Type & Questions & Tours & Average Q length & Average Q length \\
  &   &   &  in tokens & in symbols \\

\hline
Jeopardy! & 29,375 & 452 & 14.28 & 98.37 \\
ChGK Synchron & 48,065 & 1,821 & 32 & 234\\
ChGK Lite & 1,936 & 54 & 27.5 & 201 \\
\hline
All & 379,284 & 4,816 & 34 & 244.9 \\
\hline
\end{tabular}
\caption{Details about the sports ChGK database, as of 6 July 2021 and partially updated 18 November 2021. Tours -- tournaments; Q -- question.}\label{tab1}
\end{center}
\end{table*}

\begin{table}[ht]
\begin{center}
\begin{tabular}{|l|l|l|}
\hline
Part-of-speech & No. of words & \% \\
\hline
\multicolumn{3}{c}{\em Questions} \\
\hline
All nouns: & 160,844 & 62.12 \\
\hspace{0.1cm} Regular nouns & 116,945 & 45.17 (72.71)* \\
\hspace{0.1cm} Proper nouns & 43,899 & 16.95 (27.29) \\
\hspace{0.2cm} Persons** & 19,551 & 7.55 (46.61)*** \\
\hspace{0.2cm} Organizations & 12,123 & 4.68 (28.9) \\
\hspace{0.2cm} Locations & 10,268 & 3.97 (24.48) \\
Verbs & 49,671 & 19.18 \\
Adjectives & 48,407 & 18.70 \\
\textbf{Total} & \textbf{258,922} & - \\
\hline
\multicolumn{3}{c}{\em Answers} \\
\hline
All nouns: & 55,642 & 79.43 \\
\hspace{0.1cm} Regular nouns & 31,914 & 45.56 (57.36) \\
\hspace{0.1cm} Proper nouns & 23,728 & 33.87 (42.64) \\
\hspace{0.2cm} Persons & 13,579 & 19.39 (70.26) \\
\hspace{0.2cm} Organizations & 2,970 & 4.24 (15.37) \\
\hspace{0.2cm} Locations & 2,777 & 3.96 (14.37) \\
Verbs & 4,704 & 6.72 \\
Adjectives & 9,702 & 13.85 \\
\textbf{Total}  & \textbf{70,048} & - \\
\hline
\multicolumn{3}{c}{\em Topics} \\
\hline
All nouns: & 36,767 & 77.25 \\
\hspace{0.1cm} Regular nouns & 29,299 & 61.56 (79.69) \\
\hspace{0.1cm} Proper nouns & 7,468 & 15.69 (20.31) \\
Verbs & 3,411 & 7.17 \\
Adjectives & 7,415 & 15.58 \\
\textbf{Total}  & \textbf{47,593} & - \\
\hline
\end{tabular}
\caption{Distribution of parts-of-speech in questions, answers and topics of Jeopardy! data set. *For regular and proper nouns, numbers in round brackets denote percentage among all nouns. **For proper nouns, persons, organizations and locations were derived with spaCy; other entities have not been classified. ***Percentage among all defined entities.}\label{tab2}
\end{center}
\end{table}

\section{Task Discussion}

In this paragraph, we describe a challenge based on the Russian Jeopardy! data set. The challenge will be held in two formats: online and offline. The online format will be supervised by the team of the project Russian SuperGLUE~\cite{shavrina-etal-2020-russiansuperglue}. The task will appear in the project's online system around June 2022.~\footnote{\url{https://russiansuperglue.com/tasks/}} The offline format is organized as a series of Jeopardy! games at the Tyumen State University where QA systems will compete against actual players, and the first game is scheduled in July 2022. We further describe the offline format, as our team is responsible for it.

\subsection{Jeopardy! Game Format}

Following the Jeopardy! challenge of February 2011, mentioned earlier, we propose that at our event two experienced players compete against one system. At the first challenge, we will test three different systems in two rounds of five topics (categories), the first one containing easier questions and the second -- harder. Players will change after two rounds, too, i.e. each system will be competing with two new players. The classic Jeopardy! also consists of two sets of categories, probably, because more rounds would wear players out.

As we do not test QA systems' acoustic technologies, during the game each system only needs an interface for texting which will be supervised by an operator. This interface, be it a command line or graphic user interface, will be broadcast on a screen behind players. When a new question is opened and the host starts reading it, a game manager will send the question in textual form to the system's operator. When the system returns the answer, its operator will press the signalling button. If he or she does it before other two players, the operator will read the answer, and the host will evaluate it as correct or incorrect. The operator is not allowed to change the system's answer, but he or she can abstain from pressing the button.

The rest of the game rules coincide with the classic version.~\footnote{See, for example, its layout in Wikipedia \url{https://en.wikipedia.org/wiki/Jeopardy!\#Gameplay}.} Hence, the task for QA systems is to automatically answer as many questions as possible, as correctly as possible, and as fast as possible. It is advisable that systems weigh their confidence before they return the answer, so as not to lose points on nonsensical answers. However, at our first game operators have the right to not press the button and prevent such cases.

\subsection{Test Set}

Currently, there is one open access data set for the project -- the one we described in the previous paragraph. It has not been split into training and developer sets, as it is common to use Web-connection in QA systems and ChGK questions are easily found on the Web with the help of search-engines. For the online system at Russian SuperGLUE, several ChGK authors created a closed test set of 512 questions. This test set will be placed in the system to evaluate online submissions.

As for the set at the offline game, it will also be a pack of yet unpublished questions written by authors of sports ChGK. These questions will be only textual, with no visual or audio support, and with as little metaphoricity and wordplay as possible. For each set of two rounds the questions will be organized in six topics (categories) and distributed into easier and harder. After the questions will be played at our first game, they will be added to our data set as a developer set.

\subsection{Baseline}

As the baseline for our project, we suggest the open-domain question-answering model for Russian based on Wikipedia, developed by DeepPavlov project: \url{odqa.ru_odqa_infer_wiki}. The starter code and its description can be found here \url{https://docs.deeppavlov.ai/en/master/features/skills/odqa.html}. To help developers install the baseline in Google Collaboratory, we have added a Jupyter Notebook in our mentioned repository \url{https://github.com/evrog/Russian-QA-Jeopardy}. We have tested our baseline on the first 800 questions in our data set, and manually checked correctness of answers. The result is 16 correct answers, i.e. approximately 2 per 100. We tried manual rephrasing questions into a question-like form, and it helped to get some answers right, but just in a few cases. Also, we tried adding a topic to a question and it lowered the performance from 16 to 12 correct answers. Hence, in the current version DeepPavlov cannot compare to actual players and needs training and facilitation before it is ready for competition.

\subsection{Evaluation}

At the offline event the host evaluates correctness of the answer and game managers keep the score. However, for system developers we suggest that to prepare their systems they can consider the following. The evaluation stage consists of two steps: the first metric compares the answer to the correct one, the second metric calculates the system's performance. The both metrics vary across existing QA projects.

The first metric varies as correctness of the answer can be understood differently. In case of the answer to the previously discussed question, the following variants are equally possible: ``Hercule Poirot'', ``It is Hercule Poirot.'', ``Poirot'' and other versions meaningfully referring to this character and no other. As mentioned by \newcite{chen2019evaluating}, many metrics used in QA evaluation are imported from machine translation. They also note that the METEOR metric~\cite{banerjee2005meteor} is, by the result of their study, the closest to human judgments. \cite{niwattanakul2013using,thada2013comparison,rahutomo2012semantic} consider Jaccard, Dice, Cosine Similarity and similarity distance, e.g. Damerau–Levenshtein distance~\cite{levenshtein1966binary}. The SberQuAD challenge of 2017 evaluated exact matches with the gold standard and overlaps of tokens averaged over all questions~\cite{efimov2020sberquad}.

We checked several metrics on the obtained answers from our baseline. The metrics were Cosine Similarity by spaCy, Jaccard distance, Damerau–Levenshtein edit distance, METEOR by NLTK~\cite{bird2009natural}. After we calculated the metrics for 800 answers, we sorted the range and found the number of correct answers from its top. The lowest was the result of Cosine Similarity: the first answer was incorrect, and then only one correct answer before the next error which is most likely due to absence of many words, especially proper nouns and word groups, in the model. The results of the rest of the metrics are: Damerau–Levenshtein -- 5 correct answers, Jaccard -- 9, METEOR -- 14, which corresponds to conclusions by~\cite{banerjee2005meteor}. The mean between the rounded METEOR coefficient for the last correct answer in the ranged set and for the incorrect answer next to it is ${(0.238095+0.217391)/2=0.227743}$ which we, currently, propose as the minimum to automatically evaluate answers as correct.

As for the second metric, \newcite{CALIJORNESOARES2020635} enlist the usual NLP measures such as Precision, F1 as well as less known Mean Average Precision (MAP), used in Information Retrieval. However, it is also possible to modify the scoring system of Jeopardy! to evaluate developed systems. In Jeopardy!, giving a correct answer adds points, giving an incorrect answer subtracts points, and giving no answer doesn't affect the score. Also, as mentioned, questions in each topic are organized from the easiest to the hardest. Hence, the score for the question depends on its rank in the topic.

For a given answer ${a}$, its rank ${r}$ in a topic ${t}$ is the order of the question, to which the answer is given: ${r_t}$.

\begin{equation}
f(a) =
  \begin{cases}
    a=r_t       & \quad \text{if answer is correct}\\
    a=0  & \quad \text{if no answer is given}\\
    a=-r_t  & \quad \text{if answer is incorrect}
  \end{cases}
\end{equation}

The result of the system is the sum of its scores for a set of $n$ questions: $\sum_{i=1}^{n} a_i$. The evaluation code can be found in our GitHub repository. To compare to earlier versions and other systems, the score should be evaluated on test sets of the same size.

\subsection{System Description} 

After each of our offline challenges, we will publish their results and description of participating systems. We consider it vital that system description directly states whether the system searches for the answer in the Internet, as the task is obviously harder to solve without Web-connection -- how actual players do during the game. Also, system description should clearly state which already existing software and Web-services (including search engines) the system uses beside original tools.

\section{Conclusion}

The paper describes the database of the Russian professional quiz-writers, ChGK, that contains 379,284 questions, answers and other metadata, like links to sources of questions. We organized its sub-set into the first data set of 29,375 Russian Jeopardy! (``Own Game'') questions and answers. We touch upon types of ChGK tournaments and analyze several examples of Jeopardy! questions. We also describe several statistical features of the data set. Finally, we outline the Russian Jeopardy! QA challenge that will be held in summer 2022 at the Tyumen State University: we touch upon our motivation, the game format, test set, baseline and evaluation metrics.

As mentioned earlier, the Jeopardy! data set written for tournaments by professional authors is not a trivial set of questions and poses unusual tasks for the QA field. We are hopeful that our proposed challenge will bring valuable results in Russian QA, and we also plan to continue it with the logical, and not only fact-oriented, type of questions.

\section{Bibliographical References}\label{reference}

\bibliographystyle{lrec2022-bib}
\bibliography{lrec2022-example}


\end{document}